\documentclass[letterpaper, 10 pt, conference]{ieeeconf}  

\usepackage{amsmath,amssymb,epsfig}
\usepackage{setspace}
\usepackage[final]{pdfpages}
\usepackage{cancel}
\usepackage[absolute]{textpos}
\usepackage{tikz}
\usetikzlibrary{shapes,arrows,petri,topaths}
\usepackage{tikz,fullpage}
\usepackage{tkz-berge}
\usepackage{subcaption} 
\usetikzlibrary{positioning}
\usepackage{enumerate}
\usepackage{mdframed}
\usepackage{float}
\usepackage{hyperref}	
\restylefloat{table}
\usepackage[ruled,linesnumbered]{algorithm2e}
\usepackage{bm}

\IEEEoverridecommandlockouts                              

\usepackage{multicol,lipsum}
\usepackage{mathtools}

\title{\LARGE \bf
Secure Minimum Time Planning Under Environmental Uncertainty: an Extended Treatment
}

\author{Alexander Ivanov$^{1}$ and Mark Campbell$^{2}$
\thanks{ }
\thanks{$^{1}$information 
{\tt\small aii4@cornell.edu}}%
\thanks{$^{2}$information
{\tt\small mc288@cornell.edu }}%
}

\newcommand{\beq}{\begin{equation}}
\newcommand{\eeq}{\end{equation}}
\newcommand{\beqa}{\begin{eqnarray}}
\newcommand{\eeqa}{\end{eqnarray}}
\newcommand{\beqan}{\begin{eqnarray*}}
\newcommand{\eeqan}{\end{eqnarray*}}

\newcommand{\R}{{\mathbb R}}

\newcommand{\Cobst}{\ensuremath{\mathcal{C}}_{\mathrm{obst}}}
\newcommand{\Cfree}{\ensuremath{\mathcal{C}}_{\mathrm{free}}}

\newcounter{l1}
\newcounter{l2}
\newcounter{l3}
\newcommand{\bdotlist}{\begin{list}{$\bullet$}{}}
\newcommand{\bboxlist}{\begin{list}{$\Box$}{}}
\newcommand{\bbboxlist}{\begin{list}{\raisebox{.005in}{{\tiny
$\blacksquare$ \ \ }}}{}}
\newcommand{\bdashlist}{\begin{list}{$-$}{} }
\newcommand{\blist}{\begin{list}{}{} }
\newcommand{\barablist}{\begin{list}{\arabic{l1}}{\usecounter{l1}}}
\newcommand{\balphlist}{\begin{list}{(\alph{l2})}{\usecounter{l2}}}
\newcommand{\bAlphlist}{\begin{list}{\Alph{l2}.}{\usecounter{l2}}}
\newcommand{\bdiamlist}{\begin{list}{$\diamond$}{}}
\newcommand{\bromalist}{\begin{list}{(\roman{l3})}{\usecounter{l3}}}

\newcommand{\thm}[1]{\noindent \begin{theorem} #1   \end{theorem}}
\newcommand{\prop}[1]{\begin{proposition} #1 \end{proposition}}

\newcommand{\defn}[1]{\begin{definition} {\rm #1 }
\end{definition}}

\newcommand{\cor}[1]{\begin{corollary}   #1  \end{corollary}}

\renewcommand{\theequation}{\arabic{equation}}
\newtheorem{theorem}{Theorem}[section]
\newtheorem{exercise}[theorem]{Exercise}
\newtheorem{lemma}[theorem]{Lemma}
\newtheorem{proposition}[theorem]{Proposition}
\newtheorem{corollary}[theorem]{Corollary}
\newtheorem{definition}[theorem]{Definition}
\newtheorem{remark}[theorem]{Remark}
\newtheorem{example}[theorem]{Example}

\newtheorem{assumption}[theorem]{Assumption}

\begin{document}

\setlength{\abovedisplayskip}{8pt}
\setlength{\belowdisplayskip}{8pt}
\setlength{\belowcaptionskip}{-8pt}

\maketitle
\thispagestyle{empty}
\pagestyle{empty}

\newcommand{\Chatobst}{\ensuremath{\hat{\mathcal{C}}_{\mathrm{obst}}}}
\newcommand{\Chatfree}{\ensuremath{\hat{\mathcal{C}}_{\mathrm{free}}}}



\begin{abstract}
Cyber Physical Systems (CPS) are becoming ubiquitous and affect the physical world, yet security is seldom at the forefront of their design. This is especially true of robotic control algorithms which seldom consider the effect of a cyber attack on mission objectives and success. This work presents a secure optimal control algorithm in the face of a cyber attack on a robot's knowledge of the environment. This work focuses on cyber attack, but the results generalize to incomplete or outdated information of an environment. This work fuses ideas from robust control, optimal control, and sensor based planning to provide a generalization of stopping distance in 3D. The planner is implemented in simulation and its properties are analyzed. 

\end{abstract}

\section{INTRODUCTION}
Knowledge of the environment is paramount for autonomous Cyber Physical Systems (CPS) to succeed in many tasks. Therefore, a system must be able to trust given information, or else be able to account for the possibility that its knowledge of the environment is incorrect. Automation has permeated many physically consequential tasks as far ranging as self driving cars \cite{Berman2016} and nuclear centrifuge control \cite{McMillan2010}. This ubiquity has increased the need for security in CPS. Dangers posed by security flaws are exemplified by several high profile instances such as the STUXNET worm and the reported downing of a U.S. RQ-170 drone in Iran \cite{Majumdar2014}. The STUXNET worm was successful in causing significant damage to the nuclear enrichment program of Iran, while the RQ-170 event exemplifies the kinds of dangers which must be addressed if we hope to automate vehicles on a large scale. Security flaws have also been shown in modern vehicles. Checkoway \textit{et al.} showed that the control and auxiliary systems of a car can be compromised by a Man In the Middle (MIM) attack where a hacker compromises a communication network providing the vehicle with outside information  \cite{Checkoway2011}.

\begin{figure} 
	\begin{center}
		\includegraphics[scale=.4]{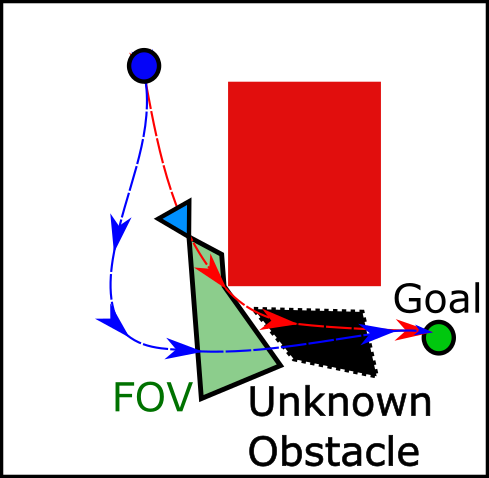}
		\caption{The robot (blue triangle) seeks to achieve the goal position (green circle). The sensor FOV is in light green, but is obstructed by the known obstacle (red) . An unsafe path (red) skirts around the obstacle. A safe (blue) path gives wide berth to the known obstacle and has time to avoid the unknown obstacle (black). \label{fig:obs_cartoon}}
	\end{center}
\end{figure}

Consider a point robot operating in a bounded, static, obstacle environment. Traditionally, knowledge of the environment, such as a map, is required and provided as a polygonal or occupancy-grid representation \cite{thrun2005probabilistic}. Many modern self driving frameworks provide this information via a networked server \cite{McMillan2010}. This work considers a scenario where the server is vulnerable to cyber attack. If a server is vulnerable, the robotic system is unsure if provided information is fully correct, incomplete, or maliciously designed. A dangerous planning scenario is exemplified by the dotted red path in Fig. \ref{fig:obs_cartoon}, where a robot is planning the fastest path around a known red obstacle, but will not be able to react in time to the unknown obstacle shown in black. If trust is to be placed in robotic vehicles, guarantees must be made on the resilience of robotic systems and planning algorithms to a variety of security threats including MIM attacks on sensors, control inputs, and assumed knowledge about the environment.

Previous work on control and planning security for CPS has had two primary components: detecting MIM attacks, and operating robotic systems under environmental uncertainty.  Mo, Chabukswar, and Sinopoli consider detection of playback attacks on a Linear Time Invariant (LTI) system, and propose a sub-optimal noisy control to increase attack detection rates. \cite{Mo2011}.  Pasqualetti \textit{et al.} provide detailed analysis of \textit{detectability} and \textit{identifiability} of noiseless LTI systems \cite{Pasqualetti2013}. Fawzi \textit{et al.} provide theoretical bounds on how large the support of an output attack vector can be while still ensuring that the attack can be detected \cite{Fawzi2014}. An NP-hard detector is also provided in the form an $l_0$ minimization, and a tractable $l_1$ approximation. Pajic \textit{et al.} utilize a similar method for noisy dynamical systems \cite{Pajic2016}. Shoukry \textit{et al.} focus on generating a \textit{sound} method using \textit{Satisfiability Modulo Theories} to ensure that a correct estimate of the attack vector is returned \cite{Shoukry2015}. 

Much work has focused on planning with uncertainty in location of impassible obstacles and robot pose. Several recent surveys detail much of the literature in this area \cite{Dadkhah2012,Hoy2015}. Much previous work focuses on Simultaneous Localization and Mapping (SLAM), which is beyond the scope of the current contribution, since this work concentrates on the planning problem. Several studies consider planning with environmental uncertainty. Missiuro and Roy adapted a Probabilistic Random Map (PRM) to account for uncertainty in a SLAM map \cite{Missiuro2006} and use linear interpolation of uncertainty ellipsoids along obstacle edges to give a probabilistic guarantee on collision avoidance. Similarly, Vitus and Tomlin study uncertainty in obstacle vertices \cite{Vitus2012a}, and provide chance constraints on obstacle avoidance. They propose a hybrid analytic-sampling method which reduces complexity compared to previous techniques. 

Current literature addresses the detection and estimation of attacks on linear systems, or considers operation with noisy sensors or adversarial pursuers. Conversely, little work has been done on planning algorithms for robots in complex environments while under malicious attack. The primary contribution of this work is an algorithm which provides guaranteed control of a robotic system under malicious attack on its knowledge of the environment. The continuous time problem is formulated and solved using direct optimization methods through a Nonlinear Program (NLP). The framework presented allows generalization of this algorithm to a variety of robotic system models.


\section{Problem Description and Definitions} \label{sect:ProbDesc}
The problem of secure control under adversarial attack is first described qualitatively. Formal definitions are then given, and a mathematical formulation is provided. The problem formulation considers a 3D Euclidean configuration space of a point robot, but the results are valid for the 2D case as demonstrated in Section \ref{sect:Sims}. 

Qualitatively, the problem statement is ``Plan a path from a start point $s$ to an end point $e$ in minimum time, and guarantee that the robot can avoid collision due to potentially malicious environmental knowledge." Several standard assumptions are required to make this problem well posed. First, a dynamically feasible path is assumed to exist between $s$ and $e$, both in the true environment as well as in the potentially malicious information. Second, the robot state at time $t$ can be represented by a point $x(t) \in \R^n$, with a sensor field of view $S_s(x) \subset \R^3$, which varies only with robot state. Finally, sensor data and control inputs are assumed un-compromised. 

In the following formulation, $J(\centerdot)$ is the cost function, which in Lagrange form is the integral of the Lagrangian $\mathcal{L}(x(t),u(t),t)$ up to the final time $t_f$. The configuration space, $\mathcal{C} = \Cobst \cup \Cfree$, is a union of the `free' space and `obstacle' or occupied space, and $u(t) \in U \subset \R^m$ are controls. A Lagrange formulation for an optimal control problem, with trusted environmental knowledge, is:

\begin{equation} \label{eq:Obj_Highlevel}
\begin{array}{clc}
\underset{u(t)\in U, t_f}{\text{minimize}} & J(x(t), u(t), t_f) & \\
s.t. & \dot{x}(t) = f(x(t),u(t))& \\
& g(x(t),u(t))\leq 0& \\
& x(t) \in \mathcal{C}_{\mathrm{free}} \quad \forall t \in (0,t_f)& \\
& x(0) = s, \quad x(t_f) = e & \\
\end{array}
\end{equation}

\noindent
In this work, the robot does not know $\Cfree$, but instead knows some unverified estimate $\Chatfree$ provided by a server. Therefore, the robot must plan over $\Chatfree$. There are many approaches to planning over an unknown space which include probabilistic approaches seeking to optimize some mean performance (expectation) as well as robust approaches which seek to find a solution for the worst-case. This work can be viewed as a form of `robust' path planning which deterministically plans over $\Chatfree$. Since $\Cfree $ is unknown, it is first prudent to consider the interaction of the FOV $S_s(x)$ and $\Cobst$. 

\defn{An obstacle $O \subset \R^3$ is \textit{visible} at time $t$ $\iff$  $O \cap S_s(x(t)) \neq \emptyset$}

The intuition behind the sequel comes from the wealth of research which has analyzed reactivity to unknown or moving obstacles in the configuration space \cite{Hoy2015,Goerzen2010}. In this work, a control law which reacts to an obstacle is called a `reactive controller'. Conceptually, when a robot encounters an unknown obstacle in the environment, new information is gained about the configuration space which may make the previously planned path infeasible or sub-optimal. Consider a locally optimal solution $x^*(t)$ to \eqref{eq:Obj_Highlevel} with the distinction that the robot plans over $\Chatfree$ instead of the true $\Cfree$. If $x^*(t) \in \Cfree$, this means that a path is safe, but perhaps sub-optimal. In other words, even if $\Chatfree$ is incorrect and some unknown obstacle $O \nsubseteq \Chatobst$ is visible at $t$, the robotic system need not react to this new information. Conversely, if $x^*(t)$ passes through such an obstacle, the robot must react. A key challenge is that the robot has dynamics and cannot stop instantaneously. Thus, if a planned path collides with an unknown obstacle, the robot must observe the obstacle far enough in advance to avoid collision. 

Note that this work does not prescribe a particular method for reacting to unknown obstacles in the environment. Instead, a reactive controller is assumed to exist with a control law, $\pi(x(t), O): \R^n \times \mathcal{M} \to U$, which is only a function of a single obstacle $O$ and $x(t)$. Here, $\mathcal{M}$ is the set of 3D non-empty polyhedra (obstacles). $\Cobst$ is a union of a finite number of such polyhedra. Furthermore, the control law $\pi(\centerdot)$ is assumed to ensure $x(t)$ remains bounded near the position where the robot first encounters an unknown obstacle, i.e. the system is stable in the sense of Lyapunov.

\begin{algorithm} 
	\KwIn{$e,s,x(0), \Chatfree$}
	\SetKwData{Xopt}{$[x^*(t),u^*(t)]$}\SetKwData{This}{this}\SetKwData{Up}{up}
	\Xopt$\leftarrow$ Solve Eq. \eqref{eq:Obj_Highlevel}\; \label{line:solve}
	\While{$\mathcal{O}_u \cup S_s(t) = \emptyset$}
	{$u(t) = u^*(t)$}\
	\If{$x^*(t) != safe$}
	{
		\While{Evasive maneuver}
		{$u(t) = \pi(x(t), O)$}
	}
	\textbf{go to} \ref{line:solve}\
	\caption{Robotic Planning Algorithm}\label{alg:LowResAlg}
\end{algorithm}
\noindent
The optimal controller solving Eq. \eqref{eq:Obj_Highlevel} is usually dubbed the path planner, while the reactive controller using $\pi(x(t), O)$ is usually called a `low level' controller. Since $\pi(x(t), O)$ is stable, it helps define a ``reactive set". Let the reactive path produced by $\pi$ be $x_r(t, x(\tau), O)$ with initial condition $x(\tau)$. 

\defn{The reactive set $S_r(x(\tau))$ defined by control law $\pi$ and the initial condition $x(\tau)$ is:
$$S_r(x(\tau)):= \underset{O \in \mathcal{M}}{\cup} x_r(t, x(\tau),O)$$}

The reactive set can be seen a generalization of stopping distance to 3D. To give some intuition to this definition, consider a one dimensional problem where a robot has dynamics $\dot{x}_1(t) = x_2(t)$, with bounded acceleration control $\dot{x}_2(t) = u(t)$, $u \in [-1,1]$. The robot's state is position and velocity. In this case, an obstacle is also a value on the real line and a reactive controller could simply try to stop the vehicle as quickly as possible: 
$$\pi(x(0), O) =  \begin{cases} 
1  & x_2(t)<0, O<x_1(\tau) \\
-1 & x_2(t)>0, O>x_1(\tau) \\
0 & o.w. 
\end{cases}$$

The reactive set is then the interval determined by the initial state and stopping distance of the robot: $S_r(x(\tau)) = [x(\tau), x(t_r)]$, where $t_r>\tau$ is the first time the robot is at rest.

\section{Planning under adversarial maps \label{sect:FullProb}}
Several arguments are necessary to motivate a solution to the problem of planning under adversarial environmental attack. First, the first argument shows that, in order to guarantee planning under adversarial attack, the reactive set $S_r(x(t))$ must be fully observed at some previous time. Second, given that $S_r(x(t))$ is guaranteed to be observed along a trajectory, one needs only ensure that $S_r(x(t)) \cap \Chatobst = \emptyset$ along a path $x(t)$ to ensure that the robot can avoid any unknown obstacle. 


Given that the robot is planning over $\Chatfree$, it is necessary to consider the effects of the mismatch between $\Chatfree$ and the true $\Cfree$ on safety. The portion of $\Cfree$ which is incorrectly assumed to be obstructed is inconsequential in terms of safety, i.e. $\Chatobst \cap \Cfree$. The `dangerous' mismatch is described by the set of unknown obstacles $\mathcal{O}_u \coloneqq \Cobst \cap \Chatfree$. When the robot senses a component of $\mathcal{O}_u$ and switches to a reactive controller, it must have a guarantee that the evasive maneuver will not cause an obstacle collision. By definition, $S_r(x(0))$ bounds all such evasive maneuvers. Since $\mathcal{O}_u$ is unknown, the only way to guarantee safety is to ensure that the robot has previously observed its planned reactive set. In other words, the following constraint must be satisfied:

\begin{equation} \label{eq:FullConstraint}
\underset{t \in [0,\tau]}{\cup} S_s(x(t)) \supseteq S_r(x(\tau))
\end{equation}

The final argument requires one more definition.

\defn{A planned path $x(t)$ defined on $t \in [0,t_f]$ is \textit{safe} if it satisfies the constraint \eqref{eq:FullConstraint} and $S_r(x(t)) \cap \Chatobst = \emptyset$ \label{defn:safe}}

Suppose that the constraint \eqref{eq:FullConstraint} is satisfied for a particular trajectory $x(t)$, and assume that the robot continuously updates its estimate $\Chatfree$ and $\Chatobst$. In practice, this implies maintaining an occupancy grid or 3D occupancy representation such as OctoMap \cite{thrun2005probabilistic}. 

\thm{Let $x^*(t)$ be a safe, planned, dynamically feasible trajectory defined on $t \in [0, t_f]$. Suppose the robot utilizes a reactive controller $\pi(x(t), O)$ and Alg.\ref{alg:LowResAlg} in Section \ref{sect:ProbDesc}. If the first unknown obstacle is seen at $\tau$, then any realized robotic path $x(t) = \{x^*(t) | t \in [0, \tau), x_r(x^*(\tau),O) \quad o.w. \}$ will be collision free. \label{thm:Safety}}

\proof{The proof will proceed by construction. Consider the first instance in time $\tau \in [0,t_f]$ when an unknown obstacle $O \subset \mathcal{O}_u$ is visible. A reactive controller will only be utilized if $O$ causes $x^*(t)$ to no longer be safe. At $\tau$, the free space estimate is updated to be 
	
$$
\Chatfree \leftarrow  \Chatfree \cup \big( \underset{t \in [0,\tau]}{\cup} S_s(x(t)) \big)/O
$$ 
and the obstacle estimate is updated to be 

$$\Chatobst \leftarrow  \big( \Chatobst \cup O \big) / \big( \underset{t \in [0,\tau]}{\cup} S_s(x(t))  \bigcap \Cfree \big).
$$

By definition, $x^*(\tau) \in S_r(x^*(\tau))$, therefore, if $x^*(t), t \in[\tau, t_f]$ remains safe under the updated $\Chatfree$, the planned trajectory can continue to be executed without re-planning and without collision. Conversely, if $x^*(t)$ is no longer safe, the control policy $\pi(x^*(\tau), O)$ is utilized. Since $O \in \mathcal{M}$ by definition, the reactive path satisfies $x_r(x^*(\tau), O) \subset S_r(x^*(\tau))$. Because constraint \eqref{eq:FullConstraint} is satisfied, and $\tau$ is the first time a component of $\mathcal{O}_u$ is visible, it must be that  $S_r(x^*(\tau)) \in \Cfree \blacksquare$.}

Theorem \eqref{thm:Safety} says that, to guarantee safety, one need only ensure the reactive sets, $S_r(x(t))$, have been observed in advance and that they do not intersect with any known obstacles: $O \in \Chatobst$. Theorem \eqref{thm:Safety} motivates the final formulation of the optimal control problem under adversarial attack:

\begin{equation} \label{eq:FullProb}
\begin{array}{clc}
\underset{u(t)\in U, t_f}{\text{minimize}} & J(x(t), u(t), t_f) & \\
s.t. & \dot{x}(t) = f(x(t),u(t))& \\
& S_r(x(t)) \subset \underset{\tau \in [0,t]}{\cup} S_s(x(\tau)) & \\
& S_r(x(t)) \in \Chatfree \quad \forall t \in (0,t_f)& \\
& x(0) = s, \quad x(t_f) = e & \\
\end{array}
\end{equation}

\section{Approximating the Set Constraints}
The set inclusion constraint in \eqref{eq:FullProb} is intuitive, but is difficult to compute in practice as it requires a union of sets which has no analytic form in general. Conversely, if the set constraints can be written as path inequality constraints of the form in \eqref{eq:Obj_Highlevel}, trajectory optimization using direct collocation methods could be used to solve the optimal control problem.

\subsection{The Visibility Constraint}
Definition \eqref{defn:safe} requires that all parts of $S_r(x(t))$ are observed by the time the robot is at $x(t)$. A tighter constraint is:

\begin{equation}
S_r(x(t)) \subset S_s(x(\tau)), \quad 0<\tau<t 
\end{equation}

\noindent
In other words, the reactive set at time $t$ must be seen, in its entirety, at some particular previous time instance. Therefore, a solution to the following optimal control problem would also be a solution to problem \eqref{eq:FullProb}:

\begin{equation} \label{eq:TightProb}
\begin{array}{clc}
\underset{u(t)\in U, t_f}{\text{minimize}} & J(x(t), u(t), t_f)& \\
s.t. & \dot{x}(t) = f(x(t),u(t))& \\
& \exists \tau<t \quad s.t. \quad S_r(x(t)) \subset S_s(x(\tau))& \\
& S_r(x(t)) \in \Chatfree \quad \forall t \in (0,t_f)& \\
& x(0) = s, \quad x(t_f) = e & \\
\end{array}
\end{equation}

\subsection{The Reactive Set} \label{sect:ReactiveSet}

To achieve a practical solution, an upper bound is proposed for the reactive set. This work uses an ellipsoidal upper bound to approximate $S_r(\centerdot)$. There are several properties which make an ellipsoidal approximation appealing. First, bounding simulated or observed trajectory data with ellipsoids is relatively easy. Second, ellipsoids are easily manipulated to provide constraint equations of the form in problem \eqref{eq:Obj_Highlevel}. Finally, if a direct collocation method is utilized to perform the discretization and optimization of the optimal control problem, the constraint $S_r(x(t)) \in \Chatfree$ can be formed into a state-varying distance condition. Details on how to practically perform this approximation are given in Section \ref{sect:Sims} and \cite{CPS_Planner_Code}. Let $Q(\centerdot)$ be a real, positive-definite, diagonal, and continuously varying matrix. Let $R(x)$, $c(x)$, and $a(x)$ be the rotation matrix, position of the robot, and ellipse center respectively. Let $z \in \R^3$. For the remainder of this work, $S_r(\centerdot)$ is assumed to take the form:

\begin{multline} \label{eq:ReactEllipse}
S_r(x) =  \\ \{y= R(x)(\sqrt{Q(x)}z+a(x)) +c(x)| z^Tz \leq 1\}
\end{multline}
\noindent
 Clearly $Q$ and $a$ depend on $x$ since the stopping distance of an inertial object depends on its initial momentum. 

\section{Discretizing the Problem} \label{sect:Disretization}

Direct trajectory optimization methods have seen wide acceptance in recent robotic literature and practice \cite{Betts2009}. Their appeal comes from recent improvements in optimization tools, their wide availably \cite{Boyd2004}, and because these tools can find locally optimal solutions even in the face of non-linear and non-convex constraints. In addition, direct methods require less expert knowledge in optimal control when compared to methods based on the calculus of variations, dynamic programming, or fast marching \cite{Betts2009}.  

The above analysis motivates the use of a direct optimization method for the problem presented in this work since the constraints proposed are complex (nonlinear, time-varying) and vary with the model of the FOV and $S_r(\centerdot)$. This section presents a Nonlinear Program (NLP) discretization of problem \eqref{eq:TightProb}. The connection between this NLP and \eqref{eq:TightProb} is not direct; additional details are provided in the appendix.  Finally, a model of the sensor set $S_s(x)$ is presented and the set inclusion constraints in \eqref{eq:TightProb} are converted into inequality constraints.  

\subsection{Nonlinear program formulation}

The formulation in \eqref{eq:TightProb} differs from standard trajectory optimization problems only in its set constraint requiring visibility and non-collision of $S_r(x(t))$. Reference  \cite{Betts2009} provides details on standard discretizations of the objective function, dynamic, and obstacle constraints. This section presents details on the set inclusion constraint. 

Let $x_i$ be the $i$th discretized state, where $x_i := x(t_i)$. Following the notation in  \cite{Betts2009}, the $i$th dynamics defect is denoted $\zeta_i$. In addition $L(\centerdot)$ is the discretized, numerically integrated form of $\mathcal{L}(\centerdot)$. The visibility constraint can be discretized by requiring that $S_r(x_i)$ is observed by a previous discretization point with a fixed time index difference:

\begin{equation} \label{eq:NLP}
\begin{array}{cll}
\underset{u_i\in U, t_f}{\text{minimize}} & L(x_1,..,x_K,u_1,..,u_K,t_f) & \\
s.t. & \zeta_i =0, d(S_r(x_i), \Chatobst) >0 & \forall i \\
& S_r(x_i) \subset S_s(x_j) \\
& x_1 = s, \quad x_K = e & \\
&j = i - \Delta i, i \in \{\Delta i,..,K\} &
\end{array}
\end{equation}

\noindent
where the distance $d(\centerdot, \centerdot)$ is Euclidean when referring to points and Hausdorff  between sets in $\R^n$.

Since $\Delta i$ is fixed and the union of sets has been ignored,  $S_r(x_i) \subset S_s(x_j)$ is a tight discrete approximation of the set inclusion constraint in Eq. \ref{eq:TightProb}. Requiring that  $\Delta i$ is fixed may be overly restrictive, and a relaxation of this constraint has been left for further study.

\subsection{Sensor model and visibility constraint}
Now that the method of discretization has been exemplified, the visibility constraint in \eqref{eq:NLP} must be converted into a set of inequalities. To do this, a model of the sensor is required.

In this work, the sensor is represented by a subset of $\R^2$ or $\R^3$, depending on the configuration space of the robot. In 2D, many sensor models such as LIDAR can be modeled as a `slice' of a disc with the radius being the effective sensor range $r$. A more conservative approximation is an isosceles triangle with equal sides of length $r$. A similar assumption is made for cameras in $\R^3$ using the pinhole-camera assumption and assuming a rectangular digital image sensor. The set $S_s(x)$ is then a polyhedron in the appropriate dimension, and can be expressed through a finite number of linear inequalities. 

\begin{equation}\label{eq:SensorModel}
S_s(x) = \{y| A(x)y \leq b(x) \} \quad y \in \R^3
\end{equation}

Since the FOV is not time dependent, $S_s(x)$ is a rotation and translation of $S_s(0)$. Let $R_{i}$ be the rotation matrix (in 2D or 3D) of the robot at time $t_i$. Given the representation of $S_r()$ and $S_s()$, the reactive set constraint in \eqref{eq:NLP} can be written concisely as: 

\begin{multline}
\label{eq:NLPElipseInclude}
\forall i \in \{\Delta i,..,K\}, \quad  j = i - \Delta i \\
A(x_j)(R_ia(x_i) + c(x_i)) \leq b(x_j) \\
d(\tilde{S}_s(x_j, x_i), B_0(1)) \geq 1
\end{multline}

\noindent
where $\tilde{S}_s(x_j, x_i)$ is $S_s(x_j)$ transformed via the bijective mapping $\mathcal{F}: S_r(x_i) \to B_0(1)$. Equation \eqref{eq:NLPElipseInclude} states that the origin must be within the translated and skewed FOV, and that the FOV's sides must be outside the unit ball $B_0(1)$. Unfortunately, the constraint is not convex in $x_i$ due to the rotation $R_i$ and other nonlinear dependence on $x_i$. Note that, in the case of polygonal obstacles the obstacle distance constraint in \eqref{eq:NLP} is computed by transforming obstacles via $\mathcal{F}$ and ensuring that the transformed obstacles' sides are outside the unit sphere.

\section{Results} \label{sect:Sims}

\begin{figure*}[t]
	\centering
	\subcaptionbox{\label{fig:ThreadNeedle1}}[.42\linewidth][c]{%
		\includegraphics[width=.42\linewidth]{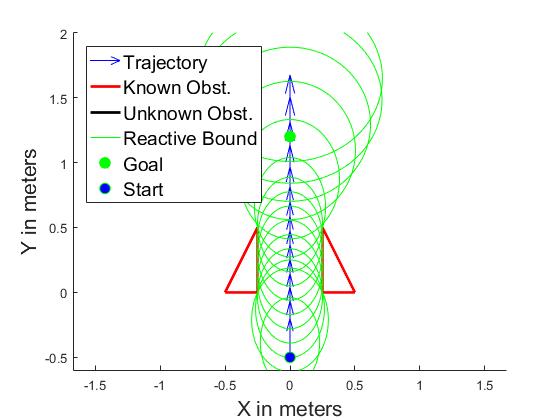}}\quad
	\subcaptionbox{ \label{fig:ThreadNeedle2}}[.42\linewidth][c]{%
		\includegraphics[width=.42\linewidth]{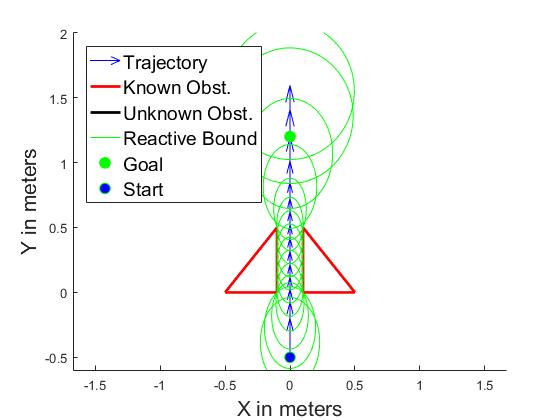}}\quad
	\caption{Slowing behavior of the secure planner.}\quad
	\label{fig:ThreadNeedle}
\end{figure*}

Several examples are used to exemplify behavior of the secure planning formulation. In the following results the same objective function, minimum time, is used. In addition, a Model Predictive Control (MPC) reactive controller is utilized by all planners and penalizes distance to the unknown obstacle and from the position $x(\tau)$. A bounded acceleration differential drive model is utilized to emulate the behavior of ground robots \cite{thrun2005probabilistic}. For details on the exact formulation of both the NLP in \eqref{eq:NLP}, the reactive controller, as well as a MATLAB simulator, see \cite{CPS_Planner_Code} and the documentation therein. Finally, timing data are presented comparing a baseline trajectory optimizer to one solving problem \eqref{eq:NLP}. 

In Figure \ref{fig:ThreadNeedle}, the robot is tasked with navigating a narrow passageway with no unknown obstacles and starts with an initial velocity of 1m/s. In Figure \ref{fig:ThreadNeedle1}, the passageway has a width of 50cm. The robot's initial velocity reduces while in the passageway  to ensure that the ellipsoidal reactive set, in green, shrinks to fit within the polygonal constraints. As the robot exits the passageway, it accelerates quickly to reduce the objective function ($t_f$). As mentioned in section \ref{sect:ReactiveSet}, $S_r(x(t))$ depends primarily on linear and angular velocity (momentum) and actuation limits. In Figure \ref{fig:ThreadNeedle2} the passageway has been narrowed to 20cm. In this case, a reduction in speed is noticeable by the smaller velocity arrows and their increased density while traversing the passageway. The robot achieves the goal in Figure \ref{fig:ThreadNeedle1} in 1.60 seconds while that in Figure \ref{fig:ThreadNeedle2} requires 2.12 seconds.

\begin{figure*}[t]
	\centering
	\subcaptionbox{Initial planned trajectory.\label{fig:BasePlanner1}}[.42\linewidth][c]{%
		\includegraphics[width=.42\linewidth]{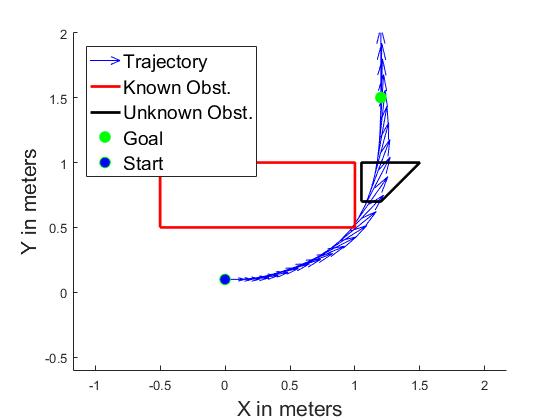}}\quad
	\subcaptionbox{Realized trajectory and collision \label{fig:BasePlanner2}}[.42\linewidth][c]{%
		\includegraphics[width=.42\linewidth]{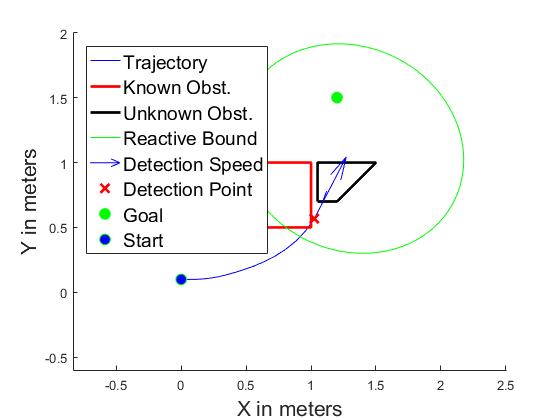}}
	\vspace{4pt}
	\caption{Behavior of a base-line minimum-time planner, with dynamics constraints. The planned trajectory takes 1.67 seconds, but results in a collision. }
	\label{fig:BasePlanner}
\end{figure*}

\begin{figure*}[t]
	\centering
	\subcaptionbox{Initial planned trajectory.\label{fig:Gplanner1}}[.42\linewidth][c]{%
		\includegraphics[width=.42\linewidth]{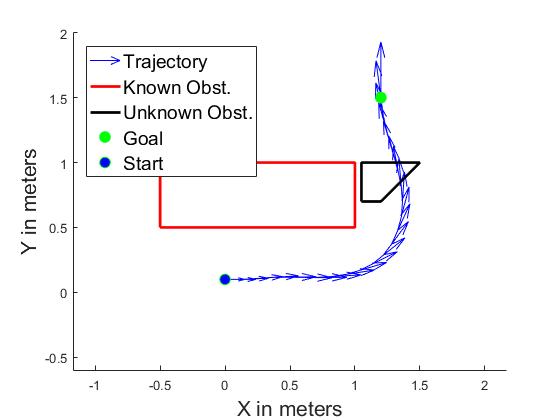}}\quad
	\subcaptionbox{Realized trajectory and avoidance. \label{fig:Gplanner2}}[.42\linewidth][c]{%
		\includegraphics[width=.42\linewidth]{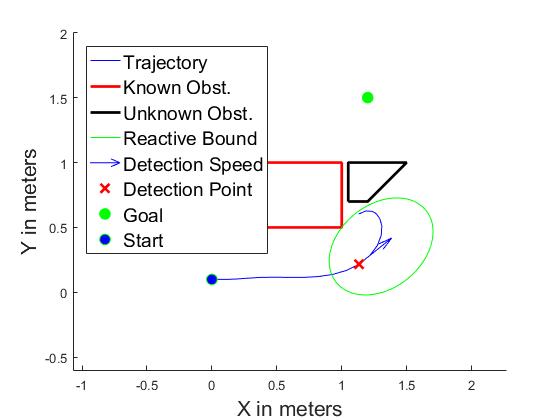}}
	\vspace{4pt}
	\caption{Behavior of the secure planner. The safe path takes 2.50 seconds.}
	\label{fig:Gplanner}
\end{figure*}

Figures  \ref{fig:BasePlanner} and \ref{fig:Gplanner} show the secure planner compared to a base-line minimum time planner with obstacle avoidance. In each case, the planners generate a minimum time trajectory; an unknown obstacle is detected as the robot rounds the corner, and the reactive controller is engaged to avoid collision. Figure \ref{fig:BasePlanner1} shows the planned trajectory of the baseline planner while Figure \ref{fig:BasePlanner2} shows the realized trajectory. To avoid visual clutter, the sensor FOV is not drawn, but has a range of $r= 2$m and angular range of $\angle{120}$. The detection point, red X, denotes the position of the robot when the unknown obstacle is observed: $x(\tau)$. In Fig. \ref{fig:BasePlanner2}, a collision results because the baseline planner did not give the reactive controller enough time to avoid the obstacle. Conversely, Fig. \ref{fig:Gplanner} shows the respective trajectories for the secure planner. The robot slows down before taking the left-hand turn in order to maintain visibility of its future reactive regions. The robot maintains a much larger distance from the known obstacle (red) and sees the unknown obstacle while initiating its turn maneuver. The reactive controller subsequently activates and successfully brings the robot to a stop. 

It is important to note that solve times depend on the initial guess provided to the optimizer, and a sufficiently poor (read infeasible) guess may cause lack of convergence. For the example in Fig. \ref{fig:Gplanner}, computation time ranged from 55-100 seconds depending on initial conditions while the base-line planner took 12-25 seconds to converge. In Fig. \ref{fig:ThreadNeedle} the baseline planner takes 1.5-3 seconds while the secure planner takes 4-6 seconds. All calculation was done in MATLAB 2016a using \textit{fmincon} and an SQP solver. Examples were run on the Windows 10 operating system and a Intel Xenon E5-1630 3.7GHz processor with no parallelization. The run-time of the guaranteed planner 5-6 times slower than the base-line planner, but the use of better optimizers, implementation in C++, and code optimization may enable near-real-time computation. In addition, the result of the base-line planner may be used as an initial guess to `warm-start' the guaranteed planner.

\section{CONCLUSIONS}
The problem of planning under adversarial attack on environmental knowledge is addressed and formulated as an optimal control problem. A novel idea of the \textit{reactive set} was introduced which generalizes stopping distance; a visibility constraint of this region is also provided. A connection was made between the continuous time-problem and that of a Mixed Integer Nonliner Program (MINP), and an asymptotic proof of safety is provided. Tightened constraints are utilized to reduce this MINP to a Nonlinear Program (NLP) enabling an efficient solution using current nonlinear optimization techniques. The reduced formulation was then implemented in simulation. 

The behavior of this formulation is analyzed through several examples. The guaranteed formulation enables a robot to act cautiously before committing to a turning maneuver around a blind corner while also trading between speed and safety. This trade prevents collisions due to uncertainty in the knowledge of the robot's environment. Although this work focuses on uncertainty stemming from a malicious attack, the techniques generalize to any cases where environmental knowledge incomplete such as areas that have not yet fully been explored.

\appendix

\subsection{The MINP Discretization}
In this appendix, the connection between the continuous time problem \eqref{eq:TightProb} and the NLP formulation is presented. As mentioned previously, we will not seek to justify standard discretizations of the objective and dynamics, but will focus on the set inclusino constraints in \eqref{eq:TightProb}.

The constraint $\exists \tau<t \quad s.t. \quad S_r(x(t)) \subset S_s(x(\tau))$, can be read as: ``The reactive set must be observed at some previous time in its entirety." Suppose a direct optimization method is employed which discretized time into $K \in \mathbb{N}^+$ intervals. The discretized visibility constraint is then:

\begin{equation}
\label{eq:MIConst}
\forall i \in \{2,..,K\} \quad  \exists j <i \quad s.t. \quad S_r(x_i) \subset S_s(x_j)
\end{equation}

\noindent
This is an integer constraint since it requires an assignment of a previous time step $j$ for each time-step $i$. Therefore, the presented discretized problem is a Mixed Integer Nonlinear Program (MINP). 

\begin{equation} \label{eq:MINP}
\begin{array}{clc}
\underset{u_i\in U, t_f}{\text{minimize}} & L(x_1,..,x_K,u_1,..,u_K,t_f) & \\
s.t. & \zeta_i =0& \\
& \forall i \in \{1,..,K\} \quad  \exists j <i \\
& s.t. \quad S_r(x_i) \subset S_s(x_j) &\\
& S_r(x_i) \in \Chatfree & \\
& x_1 = s, \quad x_K = e & \\
\end{array}
\end{equation}

\subsection{Proof of Safety}
Given the discretization in \eqref{eq:MINP}, it is desirable to show convergence properties for the integer constraint. There have been several works, such as \cite{Hargraves1987}, which show that the maximum error in the dynamics violation approaches zero as the number of discretization points $K \to \infty$ due to the defect constraints $\zeta_i$. In the case of \eqref{eq:MIConst}, it is shown that a fine enough discretization of a path which ensures that \eqref{eq:MIConst} holds, also ensures the continuous constraint is satisfied. Several smoothness assumptions are required. 

\begin{assumption} \label{asm:1}
	The reactive and sensor sets are defined respectively by Lipschitz continuous functions: $S_r: \R^n \to \mathcal{M}, S_s: \R^n \to \mathcal{M} $ with Lipschitz constants $L_r$ and $L_s$.
\end{assumption}
\begin{assumption}\label{asm:2}
	A feasible solution $x^*(t)$ to \eqref{eq:TightProb}, satisfies the set containment constraint loosely in the sense that: $ \exists \epsilon>0$ fixed $ s.t. \quad S_r(x(t),t) + B_{\epsilon}(0) \subset S_s(\tau)$
\end{assumption}

Given these assumptions:

\prop{For any path, $x^*(t)$, which satisfies assumptions \ref{asm:1}, \ref{asm:2}, there exists an integer $M \in \mathbb{N}^+$ such that $\forall K \geq M$, the the discrete set $\{x_i| x^*(i t_f/K) = x_i, i=0,1,...,K \}$ has the property that $S_r(x^*(t)) \subset S_s(x_i)$ for some $i \leq Kt/t_f$ \label{prop:Inclusion}}

\proof{During the following argument, the distance $d(\centerdot, \centerdot)$ is Euclidean when referring to points and Hausdorff  between sets in $\R^n$. Consider an arbitrary point on the path $x^*(t_0)$. Since this point is part of the solution to \eqref{eq:TightProb} and satisfies assumption \eqref{asm:2}, $\exists 0\leq t_1<t_0$ such that $S_r(x(t_0)) + B_{\epsilon}(0) \subset S_s(x(t_1)) $. By assumption \eqref{asm:1}, there exists a $\delta>0$ small enough so that $d(x_1,x_2)<\delta \implies d(S_r(x_1), S_r(x_2)) < \frac{\epsilon}{2}$, and also $d(S_s(x_1), S_s(x_2)) < \frac{\epsilon}{2}$. This comes from taking the maximum of the two functions' Lipschitz constants. 
	
Since $x^*(t)$ is continuous on the interval $t\in [0,t_f]$ it is uniformly continuous and $\Delta t>0$ can be found s.t. 

$$\forall \tau \in [-\Delta t, \Delta t], \quad d(x^*(t), x^*(t \pm \tau)) \leq \delta$$. 

In particular, this is true for $t=t_1$. Thus it follows that $S_r(x^*(t_0) \subset S_s(x^*(t_1 \pm \tau))$. It is clear that $M = \lceil t_f/\Delta t \rceil$ then satisfies the statement of the proposition. $\blacksquare$}

Prop. \eqref{prop:Inclusion} states that only a finite number of points along a trajectory can guarantee the set inclusion constraint is satisfied along an entire path. Since $\epsilon$ can be arbitrarily small, the satisfaction of the continuous constraint is well approximated by the discretization in \eqref{eq:MIConst}. Note that, if $x(t)$ and $Q(x)$ are Lipchitz, the models provided in section \ref{sect:Disretization} ensure that assumptions \ref{asm:1} and \ref{asm:2} hold. In addition, Prop. \eqref{prop:Inclusion} trivially leads to the following corollary: 

\cor{Any time discretization of an optimal path $\{ x^*(t_i)| 0<t_1<t_2,...,t_K \leq t_f \}$ such that $|t_{i+1} - t_i| \leq \Delta t$, has the property that $S_r(x^*(t)) \subset S_s(x^*(t_i))$ for some $i$ where $ t_i < t$}



\bibliographystyle{ieeetr}
\bibliography{cps_mm_2018}




\addtolength{\textheight}{-12cm}   


\end{document}